\documentclass[runningheads]{llncs}
\usepackage{graphicx}
\usepackage{amsmath,amssymb} % define this before the line numbering.
\usepackage{color}
\usepackage[width=122mm,left=12mm,paperwidth=146mm,height=193mm,top=12mm,paperheight=217mm]{geometry}
\usepackage[]{caption}

\begin{document}
\title{Efficient  Egocentric Visual Perception Combining Eye-tracking, a Software Retina and Deep Learning} 
% Replace with your title

\titlerunning{EPIC Workshop @ ECCV2018, 9 September, Munich, Germany.}
% Replace with a meaningful short version of your title
%
\author{Nina Hristozova \and Piotr Ozimek \and Jan Paul Siebert}
%
%Please write out author names in full in the paper, i.e. full given and family names. 
%If any authors have names that can be parsed into FirstName LastName in multiple ways, please include the correct parsing, in a comment to the volume editors:
%\index{Lastnames, Firstnames}
%(Do not uncomment it, because you may introduce extra index items if you do that, we will use scripts for introducing index entries...)
\authorrunning{Nina Hristozova \and Piotr Ozimek \and Jan Paul Siebert}
% Replace with shorter version of the author list. If there are more authors than fits a line, please use A. Author et al.
%

\institute{University of Glasgow, School of Computing Science, Scotland, UK. 
\email{paul.siebert@glasgow.ac.uk}\\
\url{http://www.dcs.gla.ac.uk}}
\maketitle              % typeset the header of the contribution
\begin{abstract}
We present ongoing work to harness biological approaches to achieving highly efficient egocentric perception by combining the space-variant imaging architecture of the mammalian retina with Deep Learning methods. By pre-processing images collected by means of eye-tracking glasses to control the fixation locations of a software retina model, we demonstrate that we can reduce the input to a DCNN by a factor of 3, reduce the required number of training epochs and obtain over 98\% classification rates when training and validating the system on a database of over 26,000 images of 9 object classes.

\end{abstract}
\vspace{-5mm}
\section{Introduction}
 In this paper we report the results of an experiment \cite{Hristozova} that combines  a high-resolution 50K node software retina \cite{balasuriya,OzimekDLID17,SiebertEPIC17} with a custom designed DL architecture (based on DeepFix\cite{dcnn_gaze_fixations}) coupled to an image stream collected by Tobii Pro 2 eye-tracking glasses \cite{tobiipro2_user_manual} (Figure \ref{fig:Tobii}) worn by a human observer. Our objective is to allow a human operator to collect appropriate training data for a software retina-based egocentric perception\cite{SiebertEPIC16,SiebertEPIC17} system simply by looking at objects. These objects may then be recognised in images collected by a human observer using eye tracking glasses, or a machine observer equipped with a saliency model to direct visual gaze. 

The space-variant sampling within the retina, as illustrated in Figure \ref{fig:CorticalClass}, affords almost two orders of magnitude data reduction to the brain. Furthermore, the pathway from the retina to area 1 in the visual cortex (V1) computes a form of complex-log conformal  mapping that affords a degree of scale and rotation invariance to the appearance of fixated objects\cite{OzimekDLID17,SchwartzSpatial,SchwartzComputational}. We previously reported that this mapping can reduce the input visual data rate by a factor of $_{\widetilde{~}}$$\times$7 for a 4K node retina  and $_{\widetilde{~}}$$\times$16.7 for a 50K node retina. A corresponding network size reduction of  $_{\widetilde{~}}$40\%  and  $_{\widetilde{~}}$83\% is obtained respectively, however, at the expense of an F1 classification score reduction of 0.86\% to 0.80\% as reported at \cite{OzimekDLID17,SiebertEPIC17}.

In this paper our objective is to demonstrate that we can achieve state-of-the-art recognition performance using our high-resolution retina implementation while also achieving efficiency gains. In addition, we wanted to investigate the potential to adopt a human observer for directing our software retina's  gaze to thereby construct a truly egocentric perception system suitable for both humans and robots.

\section{Approach}
\begin{figure}[h!]
\centering
\includegraphics[height=2.1cm]{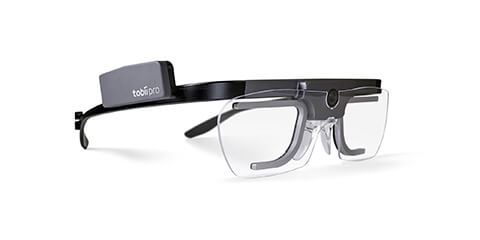}
\includegraphics[height=2.1cm]{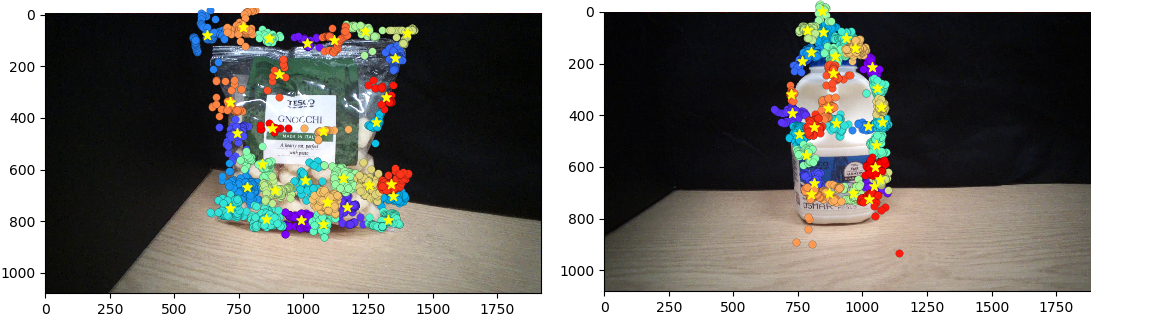}
\vspace{-1mm}
\caption{Left, Tobii Pro2 Eye Tracker Glasses; Right, Fixation clustering following homography alignment }
\label{fig:Tobii}
\vspace{-4mm}
\end{figure}

%\subsection{Pipeline}
Our processing pipeline comprises four stages: image capture, fixation cluster extraction, retina transformation and DCNN processing. Following image collection using the Tobii glasses, described below, the images for the observations of each object are composited, in order to allow the individual fixations associated with each observation (fixation) to be overlaid on a single reference image. This approximate alignment was initially achieved by means of SIFT descriptor matching and extraction of the inter-image homographies, however, we discovered that simple head stabilisation was sufficient to achieve the required image registration. \textit{K}-means clustering is then applied to these co-referenced fixation locations, where \textit{K} has been set to 1\% of the number of fixations in the observations for the current object class. This both reduces the number of fixated training images to manageable numbers and also selects locally coherent clusters of fixations, whose convex hulls are used to locate the software retina within the input image, as shown in Figure \ref{fig:Tobii}. The smaller \textit{cortical} images produced by the retina are then  input to the DCNN for both training and inference purposes.

\vspace{-6mm}

\vspace{-4mm}
\section{Data collection}
A custom interface was developed to allow an operator to control the acquisition of images using the Tobii Pro 2 eye tracking glasses. The two observers who participated in this experiment were instructed to look at locations on the surface of each object which seem particularly salient, or diagnostic of each object's identity, when collecting images. In order to obtain usable image data, it was necessary to stabilise observer's head by resting their chin on a desk surface while observing each object using the Tobii glasses and also by their consciously minimising any head movement. 

A data set of over 26,000 images was collected using the Tobii glasses, split into three categories: Training, Validation and Test. Each of these categories contains  9 object classes: Eggs, Gnocchi, Juice, Ling, Milk, Rice, Strep, VitC and Yogurt. Each of the data categories comprises the following proportion of the total data: Training 80\%, Validation 18\%  and Test  2\%. 

\vspace{-5mm}
\begin{figure}[h!]
\centering
\includegraphics[height=2cm]{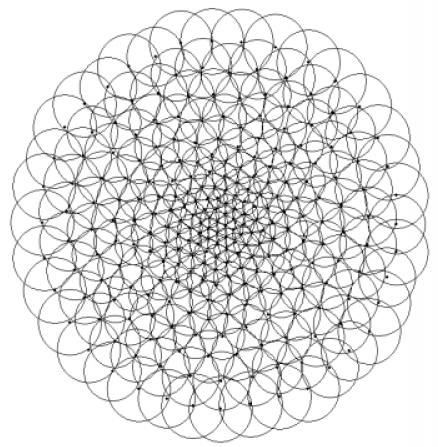}
\includegraphics[height=2cm]{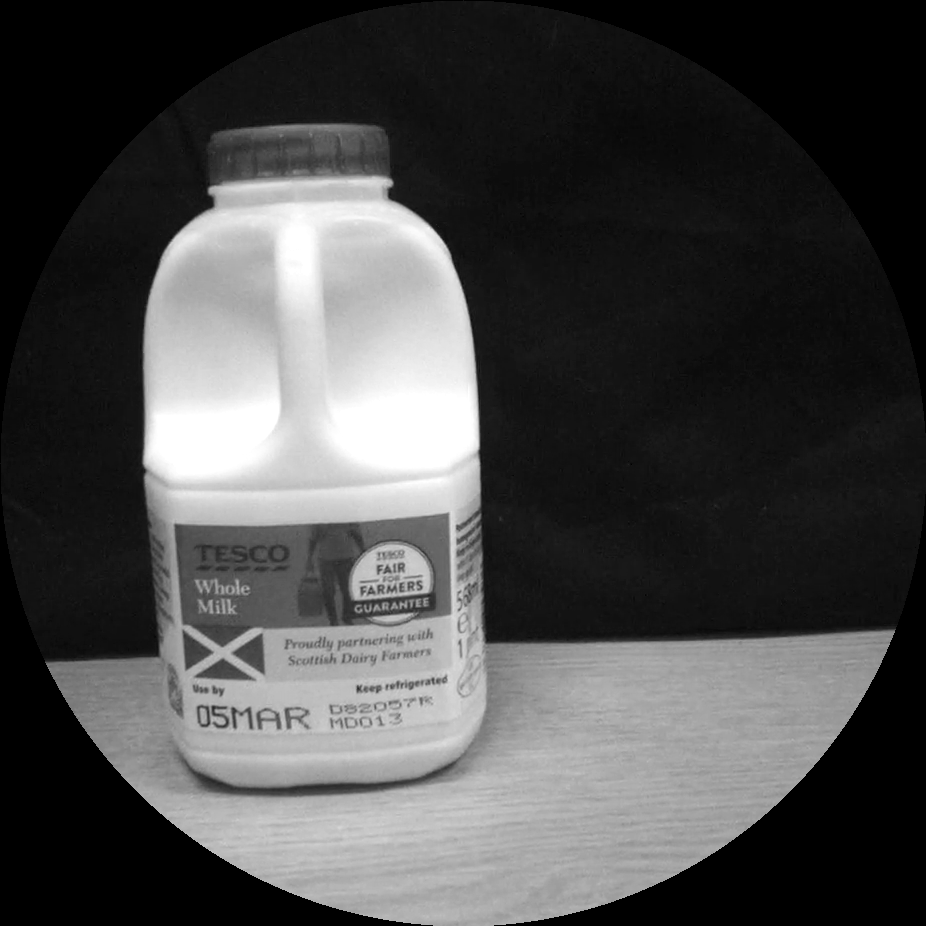}
\includegraphics[height=2cm]{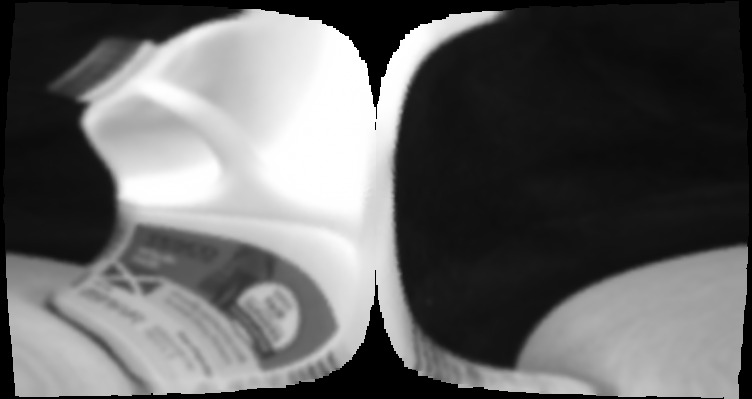}
\includegraphics[height=2cm]{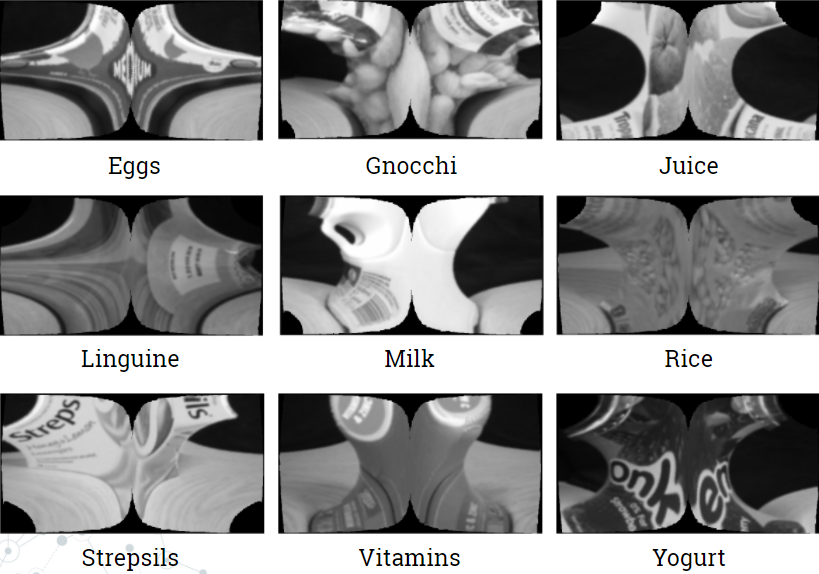}
\vspace{-2mm}
\caption{Left-Right: Retina Sampling, Fixation Crop Image, Corresponding Cortical Image, Example Cortical Images of the 9 Object Classes Captured}
\label{fig:CorticalClass}
\vspace{-5mm}
\end{figure}

\section{Building a DCNN classifier}
\label{sec:dcnns}
Inspired by Kruthiventi's DeepFix \cite{dcnn_gaze_fixations} network we developed a hand-optimised  DCNN architecture comprising seven \textit{convolutional layers(CLs)}, each followed by a max pooling layer. The input shape to the first convolutional layer was (399, 752, 3). All seven CLs have a rectifier activation function. The first two CLs have been configured with 32 convolution filters,  the second two CLs with 64 convolution filters, the fifth CL comprises 128 filters while the final two CLs comprise 256 filters. In all seven convolution layers the  filters are 3x3 in size and each of these layers is coupled by a 2x2 ReLU  max pooling function. Thereafter, the output of the last pooling layer has been flattened prior to being coupled to three \textit{fully connected(FC)} layers, each comprising 132  nodes and a final fully connected layer configured with nodes corresponding to the number of output classes, in this case 9. Each FC layer had a rectifier activation function. While the activation function of the output layer was set to \textit{softmax}, to provide accuracy and loss values ranging from 0 to 1. The classifier was optimised using \textit{stochastic gradient descent} and \textit{categorical cross entropy} was used to compute the loss function.  Drop-out was set to 50\% and only applied between the first two fully connected layers. 

\section{Cortical and Fixation Crop image DCNN validation}
\label{sebsec:classification}
In order to compare the performance obtained when pre-processing images using the software retina, as opposed to classifying standard  images, two DCNN  models were trained: the first with fixation crop images of size 926x926px and the second with the cortical images of size 399x752px. Examples of fixation crop and cortical images are given Figure \ref{fig:CorticalClass}. These images were also normalised prior to being input to the DCNN.  

\begin{figure}[h!]
\centering
\includegraphics[height=2.23cm]{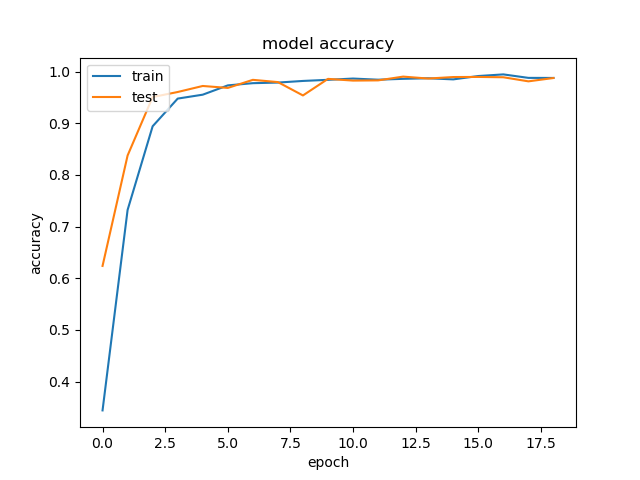}
\includegraphics[height=2.23cm]{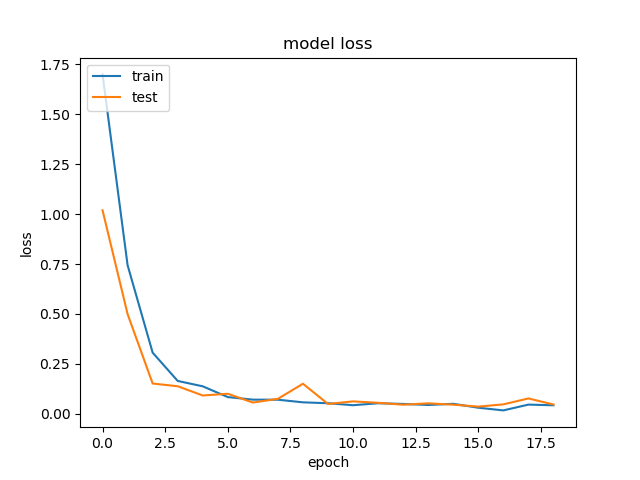}
\includegraphics[height=2.23cm]{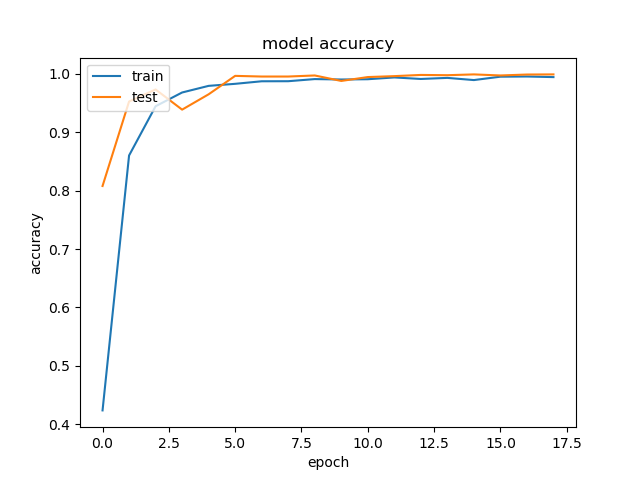}
\includegraphics[height=2.23cm]{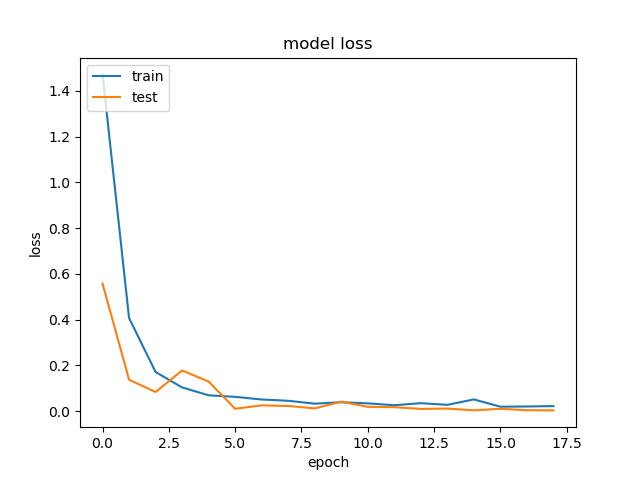}
\vspace{-2mm}
\caption{Left-Right:Cortical image accuracy, Cortical image loss, Fixation Crop image accuracy, Fixation Crop image loss, all versus training epoch.}
\label{fig:acc-f}
\vspace{-5mm}
\end{figure}

The cortical image classification DCNN model was trained using 270 steps per epoch (total number of images/batch size), where the batch size was set to 64 and the training and validation data sets comprised 21,310 and 4,800 images respectively. This model required 55 minutes processing time to execute 18 epochs and produced 98\% validation accuracy. Figure \ref{fig:acc-f} shows the accuracy and loss respectively. 6 seconds of processing time were required for this model to classify the data in its test set, resulting in an average accuracy of 98.2\%.

As illustrated in Figure \ref{fig:CorticalClass}, the fixation crop of an original image contains the retina's field of view, but retains the full image resolution. In order to benchmark the performance of the cortical image DCNN classifier, a DCNN model was trained using the full-resolution fixation crop images. In this case 1217 steps per epoch were used (again total number of images/ batch size) where the batch size is set to 16 and the training and validation data sets comprised 19,485 and  4,390 images respectively. The batch size had to be reduced to 16 from 64 used for the cortical image DCNN, because the increased numbers of pixels in the fixation crop images invoked a Tensorflow memory exhaustion error at any larger batch size. The resulting accuracy and loss are shown on Figure \ref{fig:acc-f}. This model required 2 hours and 30 minutes to execute 18 epochs and produced 99\% validation accuracy. 12 seconds of processing time were required for this model to classify the data in its test set, resulting in an average accuracy of 99.5\%.

\vspace{-5mm}

In both the cortical image and fixation crop classification experiments, the validation accuracy obtained for each classifier is close to the corresponding training accuracy result. Given the limited range of observations when the observer's head is constrained to be stationary, the captured images will be correspondingly similar. However, due to the non-linear nature of the retina transformation, the cortical images continue to exhibit significant variation in appearance.

\section{Conclusions}
In this work we have demonstrated the viability of using human fixations to drive software retina image sampling and subsequent training using our own DCNN model.  Excellent classification performance on a database of 9 object classes and a factor of 3 reduction in the size of the input network has also been obtained ($_{\widetilde{~}}$$\times$16.7 visual data reduction). However two key challenges remain: Firstly, in order to benefit fully from the data reduction afforded by the retina mapping, we must implement a custom network layer which is connected directly to the retina output samples, as opposed to generating a cortical image which is a much less efficient representation. Secondly, the current experiment required the head of the observer to be constrained in order to be able to obtain sufficiently overlapped images to undertake fixation cluster analysis. An auxiliary network is under investigation to both predict potentially diagnostic fixation points on the surface an object being learned and also to undertake (soft) object segmentation based on views of the same object taken from arbitrary and partially occluded viewpoints (e.g. when holding an object and inspecting it) to facilitate unconstrained egocentric data collection and classification by means of eye tracking glasses.

\section{Addendum}
Subsequent to preparing this paper, our recent investigations \cite{Shaikh:Thesis:2017} have now revealed that simply subsampling the cortical image can produce a network input data reduction of $\times$11.5 without compromising  performance when undertaking the eyetracking-based classification task described above. Furthermore, by adopting a scattered datapoint gridding algorithm, developed for astronomy purposes \cite{Winkel2016}, we have been able to produce cortical images which yield both a network input data reuction of $\times$10.8, and also a modest increase in classification accuracy.

%\clearpage

\bibliographystyle{splncs04}
\bibliography{refs}
\end{document}